\documentclass{article}
\usepackage{ijcai26}

\usepackage{times}
\usepackage{soul}
\usepackage{url}
\usepackage[hidelinks]{hyperref}
\usepackage[utf8]{inputenc}
\usepackage[small]{caption}
\usepackage{graphicx}
\usepackage{amsmath}
\usepackage{amsthm}
\usepackage{booktabs}
\usepackage{algorithm}
\usepackage{algorithmic}
\usepackage[switch]{lineno}
\usepackage{amssymb}
\usepackage{makecell}
\usepackage{xcolor}
\usepackage[table]{xcolor}
\usepackage{pifont}

\title{HDINO: A Concise and Efficient Open-Vocabulary Detector}

\author{
Hao Zhang$^1$
\and
Yiqun Wang$^2$\and
Qinran Lin$^3$\and
Runze Fan$^4$\and
Yong Li$^*$
\affiliations
$^1$Chongqing University\\
$^2$Chongqing University\\
$^3$Chongqing University\\
$^4$Chongqing University\\
$^*$Chongqing University
\emails
haozh416@163.com
}

\begin{document}

\maketitle

\begin{abstract}
Despite the growing interest in open-vocabulary object detection in recent years, most existing methods rely heavily on manually curated fine-grained training datasets as well as resource-intensive layer-wise cross-modal feature extraction. In this paper, we propose HDINO, a concise yet efficient open-vocabulary object detector that eliminates the dependence on these components. Specifically, we propose a two-stage training strategy built upon the transformer-based DINO model. In the first stage, noisy samples are treated as additional positive object instances to construct a One-to-Many Semantic Alignment Mechanism(O2M) between the visual and textual modalities, thereby facilitating semantic alignment. A Difficulty Weighted Classification Loss (DWCL) is also designed based on initial detection difficulty to mine hard examples and further improve model performance. In the second stage, a lightweight feature fusion module is applied to the aligned representations to enhance sensitivity to linguistic semantics. Under the Swin Transformer-T setting, HDINO-T achieves \textbf{49.2} mAP on COCO using 2.2M training images from two publicly available detection datasets, without any manual data curation and the use of grounding data, surpassing Grounding DINO-T and T-Rex2 by \textbf{0.8} mAP and \textbf{2.8} mAP, respectively, which are trained on 5.4M and 6.5M images. After fine-tuning on COCO, HDINO-T and HDINO-L further achieve \textbf{56.4} mAP and \textbf{59.2} mAP, highlighting the effectiveness and scalability of our approach. Code and models are available at \url{https://github.com/HaoZ416/HDINO}.
\end{abstract}

\section{Introduction}

Object detection aims to identify and localize objects in images, making it a crucial task in the field of computer vision. Traditional closed-set detectors~\cite{Ren1,Redmon2,Lin3,Tian4} are trained and evaluated on a fixed set of categories. Despite numerous remarkable works, the inherent nature of the real world, with its wide and diverse range of object categories, limits their practical applicability.

\begin{figure}[t]
    \centering
    \includegraphics[width=\columnwidth]{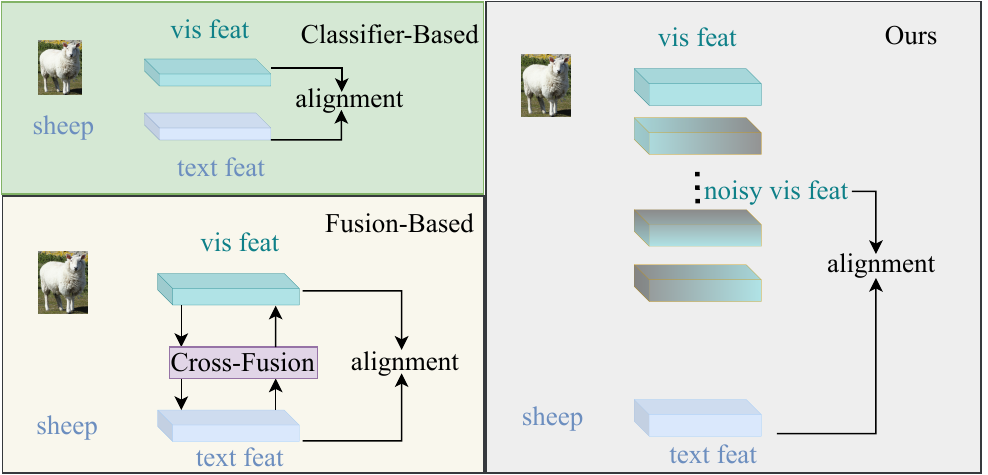}
    \caption{Comparison of different semantic alignment paradigms. Classifier-based methods align textual features only with their corresponding objects, and fusion-based methods perform alignment on global features; both exhibit limited semantic alignment capability, whereas the proposed HDINO leverages positive noisy samples to facilitate visual–textual alignment.}
    \label{fig1}
\end{figure}

\textbf{Open-vocabulary object detection (OVD)} aims to identify and localize objects that may belong to unseen categories during training, addressing the limitations of conventional closed-set detectors. Driven by the rapid progress of large language models~\cite{Jacob6} and vision-language models~\cite{Radford5,Jia7}, open-vocabulary object detectors~\cite{Zareian8,Yao9,Kamath11,Minderer14,Zhou15,Shi16,Yao17,Zhang18,Tianhe21,Zhong23} are empowered to incorporate textual knowledge acquired from massive corpora, thus being capable of detecting unseen categories during inference, which are crucial for real-world applications where object classes are diverse and constantly evolving. Among these works, a straightforward solution is to transfer semantic knowledge from vision-language models through distillation~\cite{Xiuye10,Wu24}. Additionally, to further improve cross-modal understanding, previous work~\cite{Li12,Liu13,Cheng19}incorporates fusion modules that integrate vision and language features, leading to improved detection performance. By contrast, several methods follow a relatively simple paradigm, in which text embeddings are used only as classifier prototypes and serve as contrastive objectives~\cite{Jiang20,Ao22}.

Despite achieving impressive detection performance, existing methods still suffer from two deficiencies: (1) A line of prior work incorporates textual features solely as classifier parameters~\cite{Ao22,Jiang20,Tianhe21}, which imposes strong requirements on the model architecture and the diversity of training data, due to the lack of internal semantic feature modeling within the visual representations. (2) While fusion-based methods~\cite{Li12,Liu13,Cheng19} reformulate cross-modal interactions, they typically rely on repeated integration of visual features and projected textual embeddings, incurring substantial computational overhead. Moreover, enforcing alignment between these projected text embeddings and randomly initialized visual features from scratch may compromise the integrity of the pre-trained vision-language space, leading to degraded representations. We argue that the aforementioned issues, as shown in Figure~\textcolor{red}{\ref{fig1}}, stem from a common underlying cause: inadequate optimization of semantic alignment between visual and textual modalities, thereby requiring additional architectural components or auxiliary data to compensate for the resulting performance gaps.

To this end, we present HDINO, a streamlined and high-performance open-vocabulary detector that leverages the strengths of both DINO~\cite{DINO34} and CLIP~\cite{Radford5}. Specifically, our approach is built upon a two-stage training paradigm. In the first stage, multiple noisy samples with different overlaps relative to ground-truth boxes are treated as prior positive examples and matched one-to-one with additional learnable auxiliary queries to form One-to-Many Semantic Alignment Mechanism during training. This mechanism allows the model to effectively align visual and textual features under strong prior guidance. On top of this, we introduce a Difficulty Weighted Classification Loss that emphasizes initially challenging prior samples by assigning them higher loss contributions, effectively mining hard examples and further improving overall performance. Building upon the visual–textual alignment achieved in the first stage, the second stage initializes the model with the pretrained weights and incorporates a lightweight feature fusion module, including a linear layer and a text-to-image cross-attention layer, to inject cross-modal information into the visual features and fine-tune the previous modules, thereby improving textual semantic awareness while maintaining efficiency and expressiveness.

Thanks to the above design, HDINO retains the DINO architecture at inference, except for a CLIP-based classifier and a lightweight feature fusion module. With substantially fewer parameters than Grounding DINO~\cite{Liu13}, comparable to T-Rex2~\cite{Jiang20}, and trained on roughly one-third of the publicly available data, HDINO-T reaches 49.2 mAP on COCO~\cite{COCO39}, exceeding the two baselines by 0.8 mAP and 2.8 mAP. Our contributions can be summarized as follows:
\begin{itemize}
    \item We present HDINO, a concise and efficient open-vocabulary detector that leverages DINO and CLIP to achieve strong visual–textual alignment with minimal computational overhead.
    \item We propose a two-stage training strategy: introducing a One-to-Many Semantic Alignment Mechanism in the first stage and employing a Difficulty Weighted Classification Loss to mine challenging examples, enhancing model robustness and performance. We incorporate a lightweight feature fusion module in the second stage to inject cross-modal information into visual representations while fine-tuning previously trained modules, improving textual semantic awareness.
    \item Extensive experiments show the effectiveness and efficiency of our approach with fewer parameters and training data.
\end{itemize}

\section{Related Work}
\label{sec:Related Work}

\subsection{DETR-like Models}

The pioneering DETR~\cite{Carion25} introduces the transformer architecture~\cite{transformer26} into object detection for the first time, eliminating hand-crafted components such as non-maximum suppression and anchor design. Despite its promising performance, the original DETR suffers from slow convergence, requiring long training schedules to achieve optimal performance, which makes it markedly different from traditional detectors~\cite{Ren1,Redmon2,Lin3,Tian4}. To this end, subsequent research efforts~\cite{DAB-DETR28,HDETR30,Co-DETR31,DEIM33} have been devoted to addressing this challenge. Deformable DETR~\cite{Xizhou27} enables each query to attend to a small set of keys around reference points, thereby accelerating convergence and supporting multi-scale feature representation. Alternatively, a separate line of research~\cite{DN-DETR29,DINO34} has addressed the instability inherent in the Hungarian algorithm, which can hinder training stability. In parallel, evidence indicates that the one-to-one matching scheme in DETR-like models results in sparse supervision, which poses a bottleneck to performance. Hence, subsequent works~\cite{HDETR30,Co-DETR31,Group-DETR32,DEIM33} introduce auxiliary one-to-many branches during training, following the philosophy of CNN-like detectors which perform one-to-many matching over dense feature representations, to bring more supervision signals to models. 

While our method shares the spirit of the above one-to-many approaches, it departs from them in significant ways. Group-DETR~\cite{Group-DETR32} and HDETR~\cite{HDETR30} use multiple object queries to perform one-to-many matching with the same ground-truth sample, increasing both positive and negative examples. Co-DETR~\cite{Co-DETR31} and DEIM~\cite{DEIM33}, on the other hand, rely on alternative matching strategies or data processing. Despite these differences, all these methods treat all positive samples equally, without differentiating their contributions during training. In contrast, our approach directly injects noisy samples into the original image, considers all noisy samples as positive, and employs a dedicated loss function to weight their contributions differently.

\begin{figure*}[t]
    \centering
    \includegraphics[width=\textwidth]{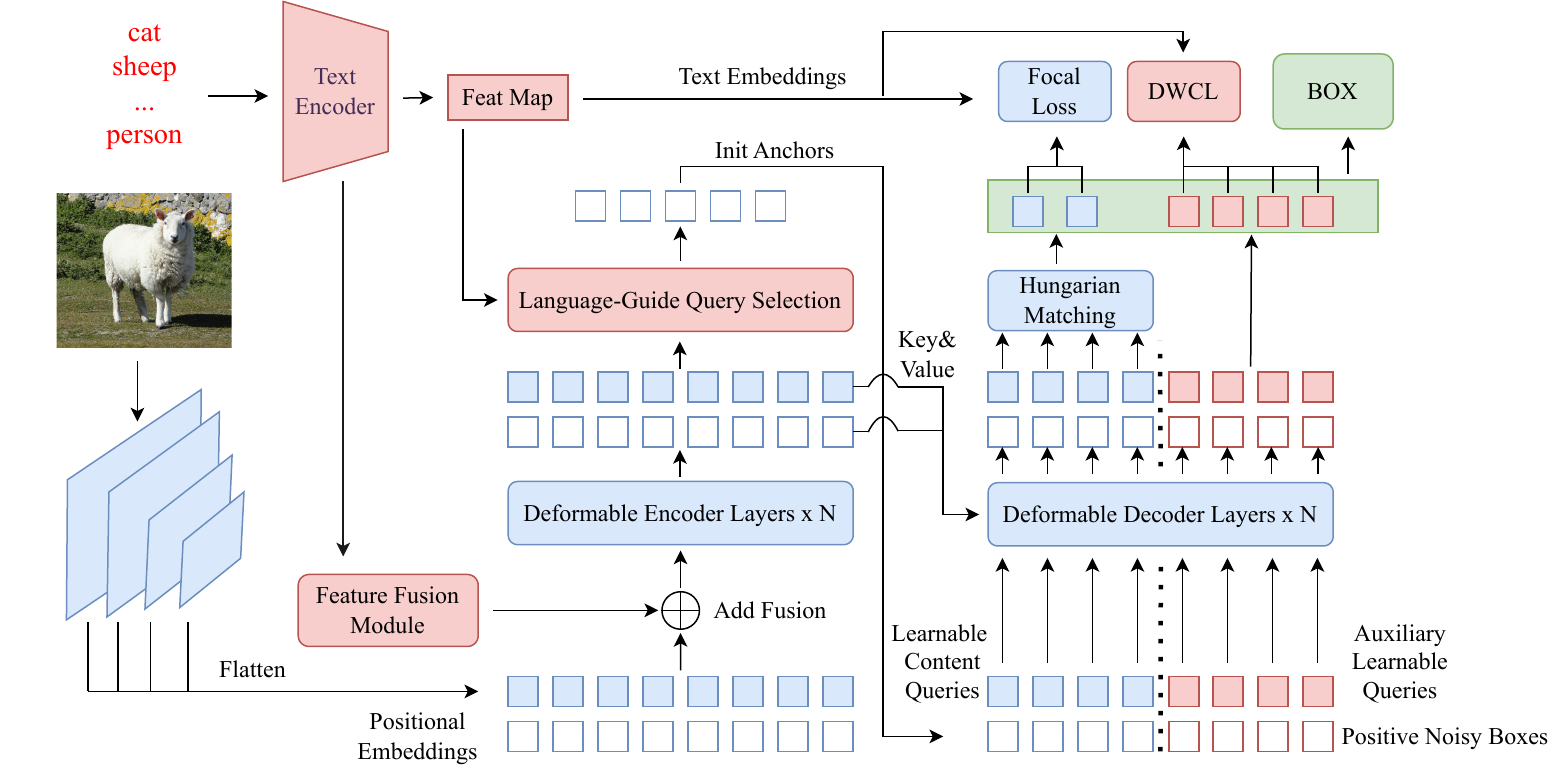}
    \caption{Overview of the HDINO model. Components in light blue are inherited from the original DINO model, modules in light red are newly introduced in HDINO, and components in light green are shared by both. The Feat Map is a linear layer that projects textual features into a unified embedding dimension, and we adopt the Language-Guided Query Selection strategy proposed in Grounding DINO to select initial anchors and remove the auxiliary queries during inference.}
    \label{fig2}
\end{figure*}

\subsection{Open-Vocabulary Detectors}

Compared with traditional closed-set detection methods, \textbf{open-vocabulary detection (OVD)}~\cite{Zareian8} aims to overcome the intrinsic fixed-category limitation of conventional counterparts, which otherwise limits their effectiveness in real-world applications. To this end, leveraging the progress in language~\cite{Jacob6} and vision-language models~\cite{Radford5,Jia7}, this line of research has emerged as a prominent direction~\cite{Zareian8,Yao9,Xiuye10,Kamath11,Minderer14,Zhou15,Shi16,Yao17,Zhang18,Zhong23,Wu24,Zang35,Zhang36}. GLIP~\cite{Li12} innovatively reformulates object detection as phrase grounding, thereby extending its training to both types of data. Grounding DINO~\cite{Liu13} advances this approach by intensifying the interaction between visual and textual modalities, thereby promoting more effective feature alignment. Apart from fusion-based models, T-Rex2~\cite{Jiang20} and YOLOE~\cite{Ao22} employ a minimalistic architecture, with text embeddings used exclusively as classifier weights. Distinct from prior approaches~\cite{Li12,Liu13,Cheng19,Ao22,Jiang20,Tianhe21}, our pipeline achieves significant performance gains with a single fusion step after high-level semantic alignment. Our results further reveal that stronger semantic alignment reduces dependence on explicit modality fusion.

\section{Method}

In this section, we present an overview of our proposed framework, as illustrated in Figure ~\textcolor{red}{\ref{fig2}}. Our method is structured in three stages: we begin with the one-to-many semantic alignment mechanism (Sec.~\ref{sec:alignment}); subsequently, we introduce the lightweight feature fusion module; finally, we provide a summary of the overall training objective (Sec.~\ref{sec:objective}).

\subsection{One-to-Many Semantic Alignment Mechanism}
\label{sec:alignment}

DETR-based models generally suffer from sparse supervision. While existing one-to-many approaches alleviate this issue, they either introduce additional negative samples or rely on auxiliary processing. In contrast, we generate only positive noisy samples from the original instances and employ a Difficulty Weighted Classification Loss to emphasize the contribution of hard examples.

\begin{figure}[t]
    \centering
    \includegraphics[width=0.8\columnwidth]{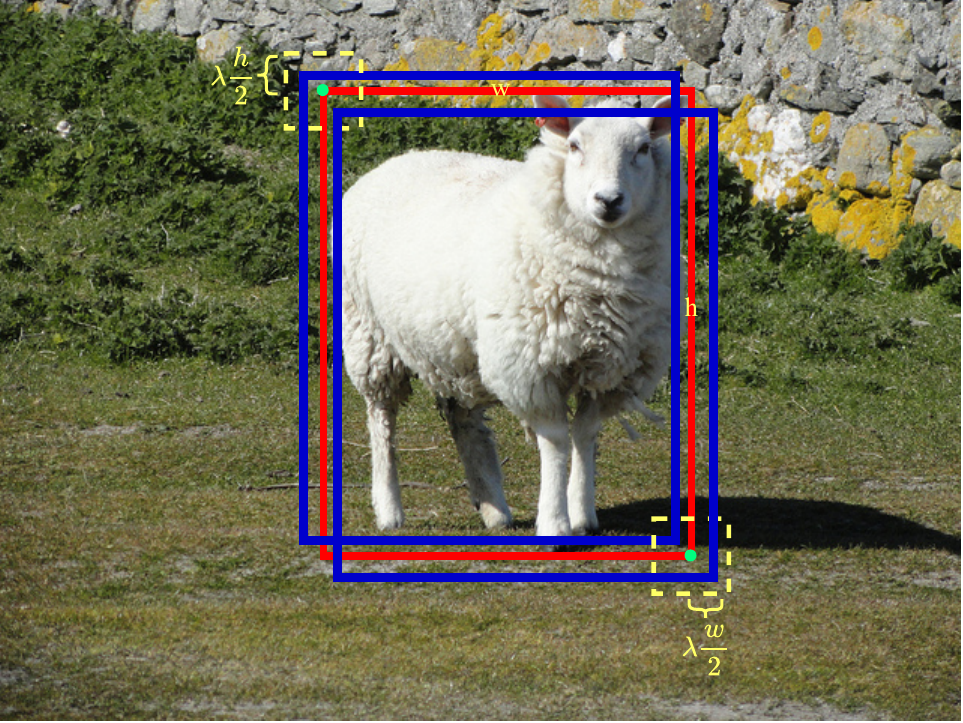}
    \caption{Visualization of positive noisy sample generation. Red boxes indicate ground-truth annotations, and blue boxes denote generated noisy samples obtained by constraining the perturbation of bounding box coordinates.}
    \label{fig3}
\end{figure}

\noindent \textbf{Noisy Positive Samples.} We generate multiple noisy boxes by randomly perturbing each ground-truth box in an image to introduce additional positive samples during training. All generated noisy boxes inherit the same category label, and corresponding instances are consistently treated as positive samples throughout the learning process, as shown in Figure~\textcolor{red}{\ref{fig3}}. Specifically, suppose an image contains $K$ ground-truth boxes. For each ground-truth box indexed by $i \in \{1, 2, \dots, K\}$ and represented as a 4D normalized vector $b_i = (x_1^i, y_1^i, x_2^i, y_2^i)$, we generate $M$ noisy boxes indexed by $j \in \{1, 2, \dots, M\}$. The width and height of the ground-truth box are computed as $w_i = x_2^i - x_1^i$ and $h_i = y_2^i - y_1^i$, respectively, which serve as reference scales for subsequent perturbations. Each noisy box $\tilde{b}_{i,j}$ is constructed by independently perturbing the top-left and bottom-right corner coordinates of $b_i$ as $(x_1^i \pm \frac{w_i}{2}\lambda \Delta x_1,\; y_1^i \pm \frac{h_i}{2}\lambda \Delta y_1,\; x_2^i \pm \frac{w_i}{2}\lambda \Delta x_2,\; y_2^i \pm \frac{h_i}{2}\lambda \Delta y_2)$, where $\Delta x_1, \Delta y_1, \Delta x_2, \Delta y_2$ are independently sampled from a standard normal distribution $\mathcal{N}(0,1)$. The scaling factor $\lambda$ controls the perturbation magnitude and is set to $0.4$ by default to ensure high-quality positive samples, i.e., those with an IoU with the corresponding ground-truth box greater than $0.5$. 

This procedure is similar to DINO; however, unlike DINO, the noisy samples in our method are used for semantic alignment rather than to stabilize early training, and no negative noisy samples are generated. Moreover, to emulate CLIP’s input images as closely as possible and to ensure that each target is surrounded by contextual background, we additionally retain $M / 3$ noisy boxes per ground-truth instance, each sharing the same center as the original box and proportionally extending all sides outward. This formulation produces positive noisy samples with varying degrees of overlap relative to the original ground-truth box, enabling the model to observe diverse positive instances under different localization difficulties while preserving semantic consistency through shared category labels.

\noindent \textbf{Auxiliary Queries.} After obtaining positive noisy samples, we introduce an additional set of $N$ learnable auxiliary queries, each of which is associated with the box information of a noisy sample and is tasked with regressing it back to the corresponding ground-truth target. From the perspective of a ground-truth object, this design assigns $M+1$ queries to the same target, including one original object query and $M$ auxiliary queries, thereby realizing a one-to-many scheme to strengthen visual–textual semantic alignment.

Notably, given that CLIP has been pretrained on large-scale image–text data, we refrain from introducing additional negative samples to avoid semantic ambiguity. Instead, all negative supervision naturally arises from mismatches between the original object queries and their predictions. To prevent information leakage, interactions from auxiliary queries to object queries are blocked in the decoder self-attention layers, and all auxiliary queries are removed at inference, keeping the inference architecture identical to that of DINO.

\noindent \textbf{Difficulty Weighted Classification Loss.} DETR-based models and their open-vocabulary extensions commonly adopt the focal loss~\cite{Lin3} as the classification objective, which is formulated as:
\begin{equation}
\mathcal{L}_{\text{focal}}(p_t) = - \alpha (1 - p_t)^{\gamma} \log(p_t)
\label{eq:focal_loss}
\end{equation}
where $\gamma$ denotes the focusing factor and $\alpha$ is the weighting factor. However, this formulation is not optimal for the auxiliary queries introduced in our framework. Specifically, the bounding boxes associated with auxiliary queries are generated by applying random perturbations sampled from a standard normal distribution to the ground-truth boxes and are therefore strictly distributed around the true object locations. Among these positive noisy samples, those with lower initial overlap with the ground-truth boxes, i.e., smaller IoU values, are inherently more difficult to classify and regress. We thus aim to design a classification loss that explicitly emphasizes hard positive samples, allowing them to contribute more to the overall loss rather than weighting samples solely by their final prediction confidence, as in the standard focal loss. 

Inspired by that larger values of $\gamma$ increase the emphasis on samples and larger values of $\alpha$ amplify their loss contribution~\cite{EFL43}, we propose Difficulty Weighted Classification Loss(DWCL) that incorporates the detection difficulty factor $1-\mathrm{IoU}$ into both focusing factor and weighting factor, enabling the loss to adaptively highlight noisy samples with higher localization difficulty:
\begin{equation}
\mathcal{L}_{\text{dwcl}}(p_l, y_l) =
\begin{cases}
- {\alpha}_{dwcl}^l (1 - p_l)^{\gamma_{dwcl}^l} \log(p_l), & \text{if } y = 1, \\
- (1 - \alpha) p_l^{\gamma} \log(1 - p_l), & \text{if } y = 0,
\end{cases}
\label{eq:focal_loss}
\end{equation}
where $l \in \{1, \dots, L\}$ denotes the index of noisy boxes, and $L$ is the batch-wise number of noisy boxes. For negative samples($y_l = 0$), the loss formulation remains identical to the standard focal loss. For positive samples ($y_l = 1$), both $\alpha_{dwcl}^l$ and $\gamma_{dwcl}^l$ are dynamically adjusted according to the detection difficulty, measured by $(1 - \mathrm{IoU})$, as defined below:
\begin{equation}
\begin{aligned}
\alpha_{dwcl}^l &= \frac{1 - \mathrm{IoU}_l}{\frac{1}{L} \sum_{l=1}^{L} \left( 1 - \mathrm{IoU}_l \right)}, \\
\gamma_{dwcl}^l &= \beta_1 \cdot (1 - \mathrm{IoU}_l) + \beta_2 .
\end{aligned}
\label{eq:difficulty_weighting}
\end{equation}
where $\beta_1$ and $\beta_2$ denote two hyperparameters.

Compared with focal loss, DWCL dynamically modulates the contribution of each noisy sample using detection difficulty as guidance, rather than treating all samples uniformly, with particular emphasis on samples that are initially farther from the ground truth and harder to detect, thereby facilitating stronger visual–textual semantic alignment.

\subsection{Feature Fusion}
\label{sec:fusion}

Prior studies have demonstrated that feature fusion can effectively enhance a model’s sensitivity to textual semantics. Motivated by this observation, we build upon the weights obtained from the first stage semantic alignment and introduce a lightweight feature fusion module to further strengthen cross-modal interactions.

Concretely, we perform feature fusion after the backbone network. A linear projection first maps textual features into a low-level visual semantic space, followed by a lightweight cross-attention layer that produces text-to-image cross-modal features. These features are then directly added to the visual representations and fed into the encoder to enhance fused representations:
\begin{equation}
\begin{gathered}
F_{\mathrm{t2i}} = \operatorname{CrossAttn}\!\left( \operatorname{Linear}(F_t),\, F_i \right), \\
F_{\mathrm{fusion}} = F_i \oplus F_{t2i}, \\
O_{\mathrm{dec}} = \operatorname{Decoder}\!\left( \operatorname{Encoder}(F_{\mathrm{fusion}}),\, Q_{\mathrm{obj}} \right)
\end{gathered}
\label{eq:fusion}
\end{equation}
where $F_i$ and $F_t$ denote visual and textual features, respectively, $F_{\mathrm{t2i}}$ denotes the text-to-image cross-modal features, $Q_{\mathrm{obj}}$ represents the object queries, and $O_{\mathrm{dec}}$ denotes the detection results.

To better preserve the semantic patterns learned in the first stage, we fine-tune the previously trained modules while adding only the feature fusion module as new parameters. By injecting cross-modal information after the backbone, visual features are encouraged to fuse with their most semantically relevant textual counterparts. The resulting fused representations are further enhanced in the encoder and naturally decoded in the decoder, without requiring explicit textual feature decoding.

\subsection{Training Objective}
\label{sec:objective}

The overall training objective comprises losses from both object and auxiliary queries. For both query types, box regression is supervised using the standard L1 loss and GIOU loss~\cite{GIOU44}. For classification, following Grounding DINO, we compute the similarity between text prompt embeddings and each detection query. The object queries are optimized with the original focal loss, while the auxiliary queries adopt the proposed Difficulty Weighted Classification Loss. Auxiliary loss after the encoder outputs and each decoder layer is used. All loss weights $\lambda_{cls}$, $\lambda_{L1}$ and $\lambda_{GIoU}$ are kept identical to those in DINO:
\begin{equation}
\begin{aligned}
\mathcal{L}_{\mathrm{obj\_query}} &= \lambda_{cls}\mathcal{L}_{\mathrm{focal}} + \lambda_{L1}\mathcal{L}_1 + \lambda_{GIoU}\mathcal{L}_{\mathrm{GIoU}}, \\
\mathcal{L}_{\mathrm{aux\_query}} &= \lambda_{cls}\mathcal{L}_{\mathrm{dwcl}} + \lambda_{L1}\mathcal{L}_1 + \lambda_{GIoU}\mathcal{L}_{\mathrm{GIoU}}, \\
\mathcal{L}_{\mathrm{total}} &= \mathcal{L}_{\mathrm{obj\_query}} + \mathcal{L}_{\mathrm{aux\_query}}
\end{aligned}
\label{eq:total_loss}
\end{equation}

\section{Experiment}
In this section, we demonstrate the effectiveness of our pipeline through extensive experimental results.\@ We train HDINO on publicly accessible datasets and conduct evaluation on COCO in a zero-shot manner. The results verify that all components consistently improve open-vocabulary object detection performance.

\subsection{Implementation Details}
\label{sec:implementation details}

\par\noindent
\textbf{Model}. HDINO is based on DINO, but discards the Contrastive DeNoising Training (CDN) module. We use  Swin Transformer~\cite{Swin-Transformer37} as the vision backbone, six deformable attention encoder layers to enhance visual representations, and six deformable attention decoder layers to decode objects. Unless otherwise noted, HDINO follows DINO's visual design. For the text encoder, a pre-trained CLIP-B~\cite{Radford5} with frozen parameters models text prompts, followed by a linear projection layer to align with a hidden dimension set to 256. 

\par\noindent
\textbf{Training.} During the entire training process, we leverage only the O365~\cite{Objects365_41} and OpenImages~\cite{openimages38} detection datasets, comprising approximately 2.2M training images. No additional data sources or other types of datasets, such as grounding datasets, are used. Following previous open-vocabulary object detection setting\cite{Li12,Liu13,Jiang20}, for each image, the ground-truth categories of the current image are used as positive prompts, whereas negative prompts are randomly sampled from the rest of the label space. All models are trained with mixed precision on eight NVIDIA L40 GPUs with a total batch size of 64. AdamW~\cite{AdamW42} is used as the optimizer, initialized with a learning rate and weight decay of $1 \times 10^{-4}$.\@ The training schedule spans about 200k iterations, where the learning rate remains $1 \times 10^{-4}$ for the first 150k iterations and is reduced by × 0.1 for the final 50k iterations. Unless stated otherwise, the remaining training configurations are consistent with those used in DINO.

\begin{table*}[t]
    \centering
    \renewcommand{\arraystretch}{1.2}
    \begin{tabular}{c c c c}
        \toprule
        Model & Backbone & Training Data & \makecell{Zero-Shot \\ 2017val} \\
        \midrule
        DyHead-T~\cite{DynamicHead45} & Swin-T & O365 & 43.6 \\
        GLIP-T (B) & Swin-T& O365& 44.9 \\
        GLIP-L & Swin-L & FourODs,GoldG,Cap24M& 49.8 \\
        Grounding-DINO-T & Swin-T & O365,GoldG,Cap4M & 48.4\\
        Grounding-DINO-L & Swin-L & O365,OpenImages,GoldG& 52.5\\
        T-Rex2-T&Swin-T&O365, GoldG, OpenImages, Bamboo, CC3M, LAION&46.4\\
        YOLO-World-S&YOLOv8-S&O365,GoldG&37.6\\
        YOLO-World-M&YOLOv8-M&O365,GoldG&42.8\\
        YOLO-World-L&YOLOv8-L&O365,GoldG&44.4\\
        YOLO-World-L&YOLOv8-L&O365,GoldG,CC3M&45.1\\
        \rowcolor{gray!15}
        HDINO-T(ours)&Swin-T&O365,OpenImages&49.2\\
        \rowcolor{gray!15}
        HDINO-L(ours)&Swin-L&O365,OpenImages&51.7\\

        \bottomrule
    \end{tabular}
    \caption{Zero-shot object detection performance comparison on COCO. All methods are evaluated on the COCO \texttt{val2017} split under the zero-shot setting. For fair comparison, we report the standard mean Average Precision (mAP) metric. HDINO is trained using only two publicly available detection datasets.}
    \label{table1}
\end{table*}

\par\noindent
\textbf{Inference}. During inference, we follow the standard zero-shot protocol and use the category names of the entire benchmark as text prompts. In addition, the auxiliary queries module can be removed, leaving all components of HDINO consistent with the DINO model. The only differences are the incorporation of a text encoder and a lightweight feature fusion module, without altering the original visual architecture.

\noindent \textbf{Zero-Shot Evaluation.} In this study, we evaluate the proposed HDINO on COCO, containing 80 categories. All evaluations are conducted in a zero-shot manner. Here, zero-shot means the model has no access to the evaluation datasets during training and must rely solely on text prompts to recognize categories. This setup strictly evaluates open-vocabulary generalization. We follow standard evaluation protocols that report the AP metric on the COCO \texttt{val2017} subset.

\subsection{Zero-Shot Detection Results}
\label{sec:results}

In this subsection, we evaluate HDINO's zero-shot detection performance against a range of state-of-the-art approaches. As summarized in Table~\textcolor{red}{\ref{table1}}, the majority of existing methods rely on heterogeneous training data, typically combining detection and grounding datasets, with several approaches being trained on over 5M images in total. By contrast, HDINO is trained solely on detection data, without introducing additional dataset types such as grounding data, and avoids computationally intensive layer-wise cross-modal feature extraction. Despite this simplified and more efficient training setup, under the Swin-T backbone, HDINO attains 49.2 mAP on the COCO benchmark, surpassing DINO-based counterparts Grounding DINO and T-Rex2 by 0.8 mAP and 2.8 mAP, respectively. Moreover, HDINO also consistently outperforms methods trained with more diverse data sources.
\begin{table}[t]
    \centering
    \renewcommand{\arraystretch}{1.2}
    \begin{tabular}{c c c c c}
    \toprule
    Model & O2M & DWCL & Fusion & AP \\
    \midrule
    DINO+CLIP & \ding{55} & \ding{55} & \ding{55} & 46.3 \\
    DINO+CLIP & \ding{51} & \ding{55} & \ding{55} & 48.3 \\
    DINO+CLIP & \ding{51} & \ding{51} & \ding{55} & 48.8 \\
    DINO+CLIP & \ding{51} & \ding{51} & \ding{51} & 49.2 \\
    \bottomrule
    \end{tabular}
    \caption{Ablation study of (i) the one-to-many semantic alignment mechanism(O2M), (ii) DWCL, and (iii) feature fusion(Fusion), as well as their full combination.
}
    \label{table2}
\end{table}

\subsection{Ablation Study}
\label{sec:ablation}

In this subsection, we perform ablation studies on the COCO dataset to systematically analyze the impact of each component and demonstrate the effectiveness of our method. All results are reported in Table~\textcolor{red}{\ref{table2}}, with all models adopting the Swin-T configuration. The ablation results highlight that the one-to-many semantic alignment mechanism is the primary factor driving the performance gains of HDINO, contributing a 2.0 mAP improvement over the DINO+CLIP baseline. This finding indicates that the standard one-to-one matching paradigm in DINO limits the ability of visual representations to effectively internalize semantic knowledge from pretrained text embeddings. As a result, simply using text features as classifier weights is insufficient for achieving strong vision–language alignment. 
\begin{table}[t]
    \centering
    \renewcommand{\arraystretch}{1.2}
    \begin{tabular}{c c c c}
    \toprule
    Loss & $\beta_1$ & $\beta_2$ & AP \\
    \midrule
    Focal Loss & \ding{55} & \ding{55} & 48.3 \\
    DWCL & 1 & 1.5 & 48.6 \\
    DWCL & 1 & 2 & \textbf{48.8} \\
    DWCL & 1 & 2.5 & 48.7 \\
    DWCL & 2 & 1 & 48.5 \\
    DWCL & 2 & 1.5 & 48.6 \\
    DWCL & 2 & 2 & 48.6 \\
    \bottomrule
    \end{tabular}
    \caption{Impact of different $\beta_1$ and $\beta_2$ hyperparameters in DWCL. All models are trained with identical pre-training configurations and evaluated on COCO for performance comparison.}
    \label{table3}
\end{table}

\begin{figure}[t]
    \centering
    \includegraphics[width=\columnwidth]{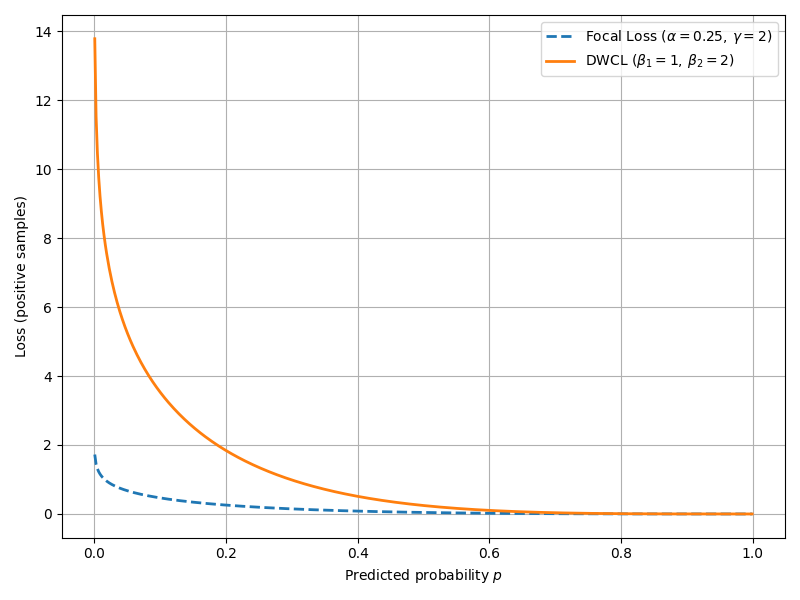}
    \caption{Comparison between the standard focal loss and the proposed Difficulty Weighted Classification Loss (DWCL) for positive samples. The loss is plotted as a function of the predicted probability $p$. For focal loss, $\alpha=0.25$ and $\gamma=2$ are used. For DWCL, the detection difficulty is fixed to $\mathrm{IoU}=0.5$ with $\beta_1=1$ and $\beta_2=2$, and the weighting factor $\alpha_{\mathrm{dwcl}}$ is normalized by $\mathbb{E}[1-\mathrm{IoU}]=0.25$.}

\label{fig4}
\end{figure}

\begin{figure}[h!]
    \centering
    \includegraphics[width=\columnwidth]{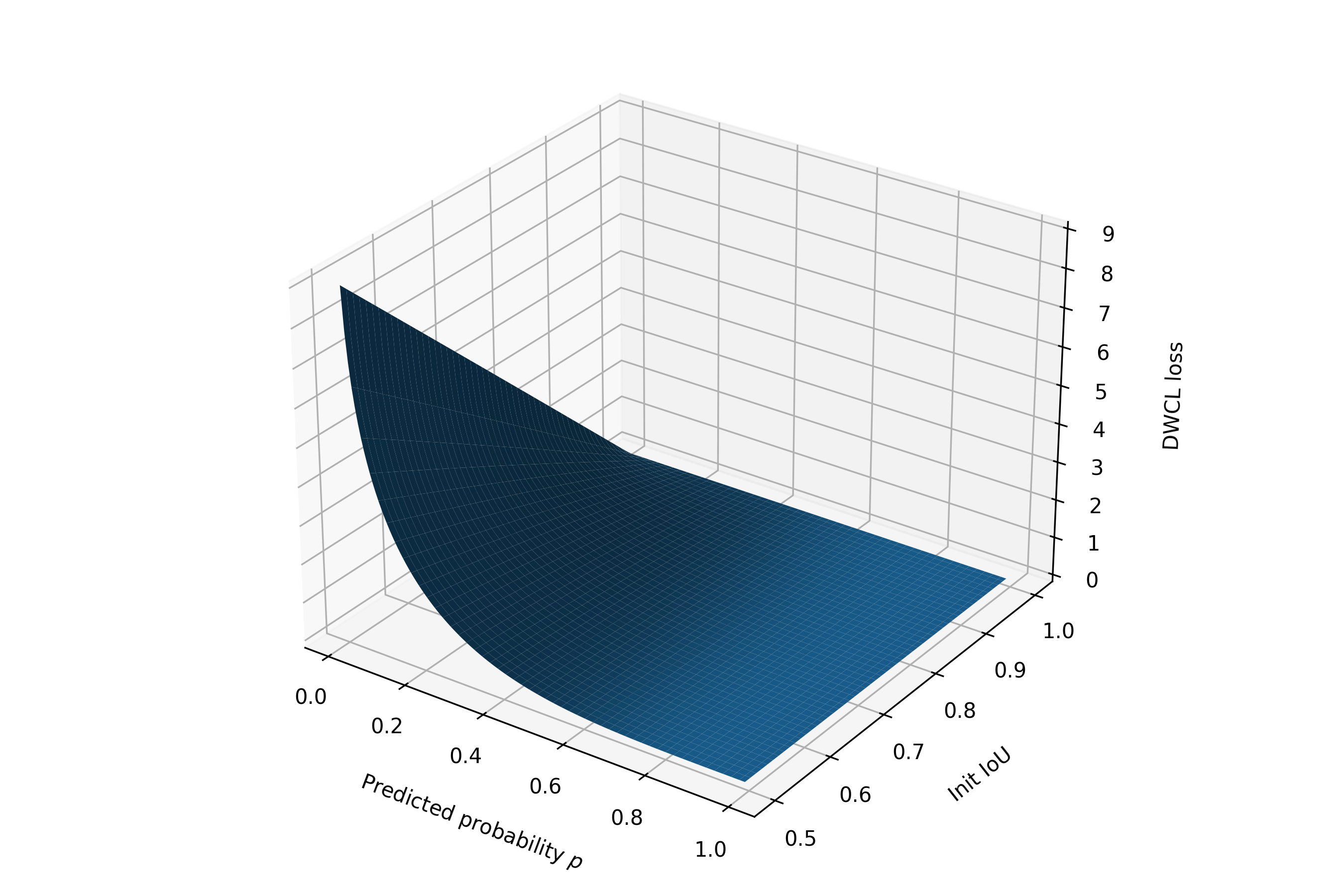}
    \caption{3D visualization of the proposed Difficulty Weighted Classification Loss (DWCL) for positive samples. In this visualization, hyperparameters are fixed to $\beta_1=1$ and $\beta_2=2$, and $\alpha_{\mathrm{dwcl}}$ is normalized by $\mathbb{E}[1-\mathrm{IoU}]=0.25$.}

\label{fig5}
\end{figure}

By relaxing this constraint through one-to-many alignment, the proposed method enables more comprehensive semantic supervision from text features, thereby substantially improving alignment quality without relying on additional data sources.
Moreover, DWCL leverages strong priors from auxiliary queries to emphasize hard examples, resulting in further gains in recognition performance. Finally, feature fusion conducted on semantically aligned representations enhances the model’s sensitivity to textual semantics, thereby reinforcing HDINO's overall text-aware detection capability. Importantly, these improvements are achieved without incorporating grounding data or computationally intensive cross-modal architectures, demonstrating that HDINO can attain competitive performance in a strictly constrained and efficient training setting.

Under the same experimental settings described above, we further investigate the impact of the hyperparameters $\beta_1$ and $\beta_2$ in DWCL, as reported in Table~\textcolor{red}{\ref{table3}}. The results show that DWCL is not highly sensitive to the specific choices of these two hyperparameters, with only marginal performance variations observed across different settings. Based on these results, we adopt the best-performing configuration, $\beta_1= 1$ and $\beta_2= 2$, for all subsequent experiments. Figure~\textcolor{red}{\ref{fig4}} compares DWCL with the standard focal loss under a representative hard positive setting. Compared to focal loss, DWCL assigns substantially larger loss values to low-confidence positive samples, indicating stronger supervision for noisy object queries with higher localization uncertainty. Meanwhile, both losses rapidly diminish for well-classified samples, showing that DWCL does not over-penalize easy positives. Figure~\textcolor{red}{\ref{fig5}} illustrates that DWCL explicitly increases the optimization focus on positive samples that are difficult to localize at early stages, thereby encouraging robust semantic alignment.

\subsection{Downstream Transferring}
\label{sec:transferring}

We transfer the pre-trained HDINO model to COCO for downstream closed-set object detection. As shown in Table~\textcolor{red}{\ref{table4}}, HDINO demonstrates strong generalization performance with only a small number of fine-tuning epochs. Following YOLOE~\cite{Ao22}, we adopt two fine-tuning strategies: (1) linear probing, where only the Feat Map classification layer is trained, and (2) full tuning, where all model parameters are updated. For linear probing, we fine-tune the model for only 10 epochs, while for full tuning, we train for 15 epochs. Remarkably, even under the linear probing setting, HDINO significantly outperforms all variants of YOLOE and YOLO-World~\cite{Cheng19} under full tuning. Under full tuning, HDINO-T and HDINO-L achieve 56.4 mAP and 59.2 mAP, respectively. These results indicate that HDINO exhibits strong generalization capability and serves as an effective pre-trained model that can be efficiently transferred to downstream tasks with minimal tuning.

\begin{table}[t]
\centering
\renewcommand{\arraystretch}{1.2}

\begin{tabular}{l c c c c}
\hline
Method & Epochs& AP$^{b}$ & AP$^{b_{50}}$ & AP$^{b_{75}}$ \\
\hline

\multicolumn{4}{c}{\emph{Linear Probing}} \\
YOLOE-v8-S & 10 & 35.6 & 51.5 & 38.9 \\
YOLOE-v8-M & 10 & 42.2 & 59.2 & 46.3 \\
YOLOE-v8-L & 10 & 45.4 & 63.3 & 50.0 \\
YOLOE-v11-S & 10 & 37.0 & 52.9 & 40.4 \\
YOLOE-v11-M & 10 & 43.1 & 60.6 & 47.4 \\
YOLOE-v11-L & 10 & 45.1 & 62.8 & 49.5 \\
\rowcolor{gray!15}
HDINO-T     & 10 & \textbf{50.7} & 65.5 & 56.0 \\
\rowcolor{gray!15}
HDINO-L      & 10 & \textbf{53.4} & 69.1 & 59.1 \\

\hline
\multicolumn{4}{c}{\emph{Full Tuning}} \\
YOLOE-v8-S & 160 & 45.0 & 61.6 & 49.1 \\
YOLOE-v8-M & 80 & 50.4 & 67.0 & 55.2 \\
YOLOE-v8-L & 80 & 53.0 & 69.8 & 57.9 \\
YOLOE-v11-S & 160 & 46.2 & 62.9 & 50.0 \\
YOLOE-v11-M & 80 & 51.3 & 68.3 & 56.0 \\
YOLOE-v11-L & 80 & 52.6 & 69.7 & 57.5 \\
YOLO-World-S & 80 & 45.9 & 62.3 & 50.1 \\
YOLO-World-M & 80 & 51.2 & 68.1 & 55.9 \\
YOLO-World-L & 80 & 53.3 & 70.1 & 58.2 \\
\rowcolor{gray!15}
HDINO-T      & 15 & \textbf{56.4} & 72.1 & 62.2 \\
\rowcolor{gray!15}
HDINO-L      & 15 & \textbf{59.2} & 75.5 & 65.5 \\
\hline
\end{tabular}
\caption{Fine-tuning results on COCO.}
\label{table4}
\end{table}

\subsection{Limitations}
\label{sec:limitations}

HDINO is pre-trained solely on detection data and demonstrates strong modality alignment ability at inference time with almost no reliance on additional modules. Nevertheless, since the goal of HDINO is to investigate the feasibility of exploiting the model’s intrinsic capacity for semantic alignment, we deliberately do not incorporate grounding data and prompt templates during training. As a result, HDINO exhibits suboptimal performance on long-tailed datasets. Addressing this limitation by integrating grounding supervision or more effective prompting strategies is left for future work.

\section{Conclusion}
\label{sec:conclusion}
In this paper, we propose HDINO, an open-vocabulary object detector trained through a two-stage pre-training scheme. In the first stage, we construct a One-to-Many Semantic Alignment Mechanism by combining positive noisy samples with the original object queries of DINO, and introduce a Difficulty Weighted Classification Loss to emphasize hard samples that are distant from the ground-truth targets at initialization. In the second stage, we incorporate a lightweight feature fusion module to enhance textual semantic awareness. These designs jointly enable HDINO to serve as a concise and efficient framework for open-vocabulary object detection.

\bibliographystyle{named}

\end{document}